\def\year{2020}\relax
\documentclass[letterpaper]{article} 
\usepackage{aaai20}  
\usepackage{times}  
\usepackage{helvet} 
\usepackage{courier}  
\usepackage[hyphens]{url}  
\usepackage{graphicx} 
\usepackage{graphicx}  
\usepackage{color}
\usepackage{cases}
\usepackage{multirow}
\usepackage{enumerate}

\usepackage{algorithm}
\usepackage{algorithmic}

\title{Asymmetric Co-Teaching for Unsupervised Cross-Domain \\ Person Re-Identification}

\author{
Fengxiang Yang,\textsuperscript{\rm 1}\thanks{This work was done when Fengxiang Yang was an intern at Youtu Lab (yangfx@stu.xmu.edu.cn).}
Ke Li,\textsuperscript{\rm 3}
Zhun Zhong,\textsuperscript{\rm 1}
Zhiming Luo,\textsuperscript{\rm 2}\thanks{Corresponding Author (zhiming.luo@xmu.edu.cn, winfredsun@tencent.com)}
Xing Sun,\textsuperscript{\rm 3}\textsuperscript{$\dag$}\\
\Large \textbf{Hao Cheng,\textsuperscript{\rm 3}
Xiaowei Guo,\textsuperscript{\rm 3}
Feiyue Huang,\textsuperscript{\rm 3}
Rongrong Ji,\textsuperscript{\rm 1}
Shaozi Li\textsuperscript{\rm 1}}\\
\textsuperscript{\rm 1}Artificial Intelligence Department, Xiamen University, China\\
\textsuperscript{\rm 2}Post Doctoral Mobile Station of Information and Communication Engineering, Xiamen University, China\\
\textsuperscript{\rm 3}Tencent Youtu Lab, Shanghai, China
}

\begin{document}
\maketitle

\begin{abstract}

Person re-identification (re-ID), is a challenging task due to the high variance within identity samples and imaging conditions. Although recent advances in deep learning have achieved remarkable accuracy in settled scenes, \textit{i.e.}, source domain, few works can generalize well on the unseen target domain. One popular solution is assigning unlabeled target images with pseudo labels by clustering, and then retraining the model. However, clustering methods tend to introduce noisy labels and discard low confidence samples as outliers, which may hinder the retraining process and thus limit the generalization ability. In this study, we argue that by explicitly adding a sample filtering procedure after the clustering, the mined examples can be much more efficiently used. To this end, we design an asymmetric co-teaching framework, which resists noisy labels by cooperating two models to select data with possibly clean labels for each other. Meanwhile, one of the models receives samples as pure as possible, while the other takes in samples as diverse as possible. This procedure encourages that the selected training samples can be both clean and miscellaneous, and that the two models can promote each other iteratively. Extensive experiments show that the proposed framework can consistently benefit most clustering based methods, and boost the state-of-the-art adaptation accuracy. Our code is available at https://github.com/FlyingRoastDuck/ACT\_AAAI20.

\end{abstract}

\section{Introduction}
\label{sec:da}

\begin{figure*}[!t]
  \includegraphics[width=\linewidth]{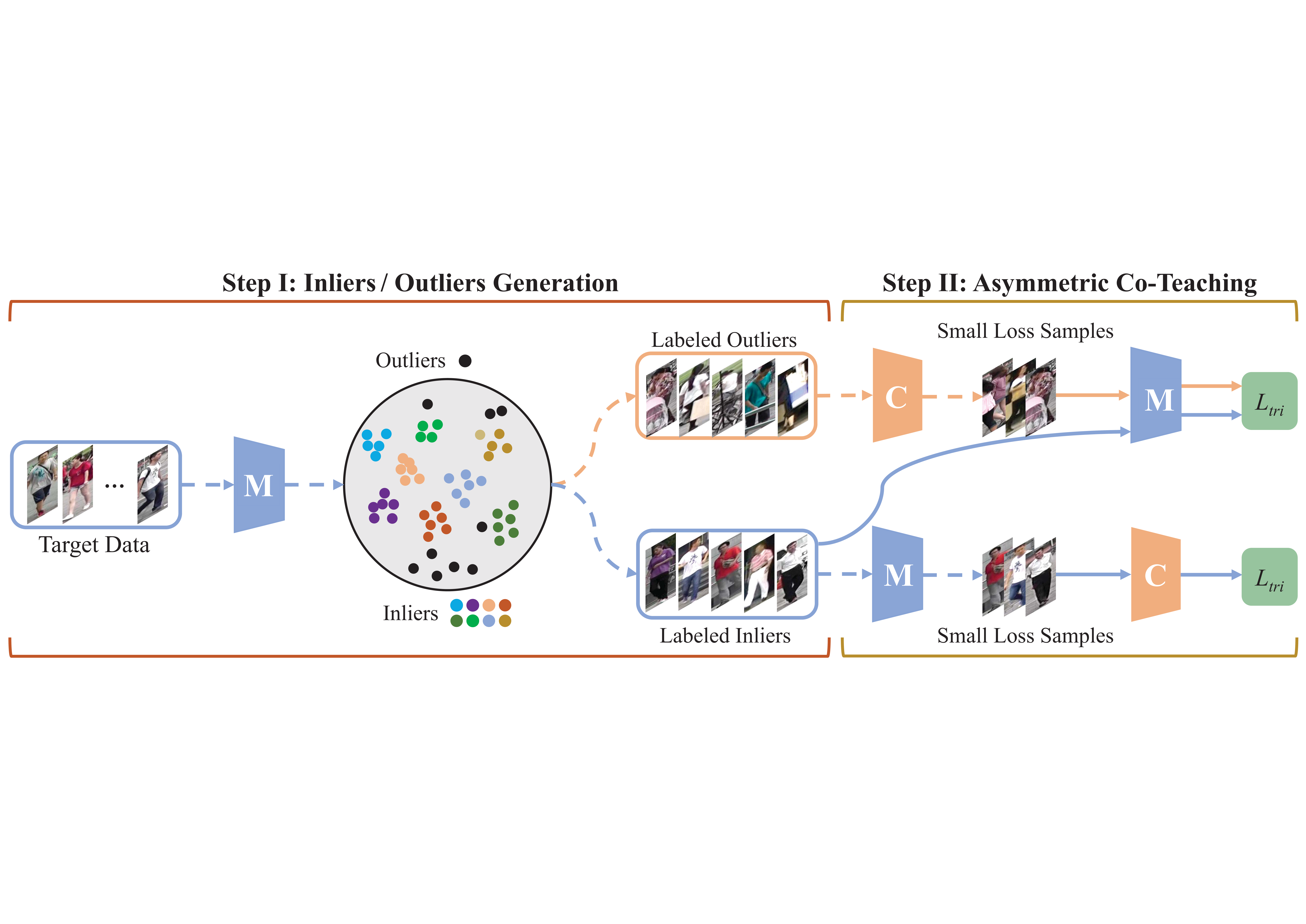}
  \label{fig:process}
  \caption{The proposed asymmetric co-teaching framework (ACT). ``M" and ``C" denote the main model and the collaborator model, respectively. We first train CNN on the source labeled data and fine-tune it on target data with pseudo labels predicted by clustering to get initial weights for ``M" and ``C". ``M" receives samples as diverse as possible from inliers and outliers, while ``C" takes in samples as pure as possible from inliers during ACT. This process encourages the two models to mutually promote the discriminative ability of each other. More details can be found at Sec.~\ref{sec:ACT}.}
\end{figure*}

Person re-identification (re-ID) \cite{sun2018beyond,zheng2016person,li2018harmonious} aims to locate the target person in surveillance videos with a given probe image. With the rapid evolution of deep learning models, the accuracy of person re-ID has been greatly boosted in the public datasets. However, models trained on the source domain often suffer from domain shifts, leading to a performance decline on a different target domain.

To alleviate this issue, recent works \cite{zhong2019learning,zhong2018camstyle} make efforts on the unsupervised domain adaptation (UDA), which aims to transfer the knowledge from the labeled source domain to the unlabeled target domain. These works mainly lie in two aspects, distribution aligning \cite{PTGAN,SPGAN,chang2019disjoint,MMFA,TJAIDL} and target pseudo label discovering \cite{fan2017unsupervised,song2018unsupervised,li2018unsupervised}. The former aims to reduce the distribution gap between domains in a common space, such as image-level \cite{PTGAN,SPGAN} and attribute-level \cite{chang2019disjoint,MMFA,TJAIDL} spaces. The latter attempts to leverage the underlying relations among target samples and predict pseudo labels for model retraining, \textit{e.g.} assigning pseudo labels based on clustering \cite{fan2017unsupervised,song2018unsupervised,li2018unsupervised} and $k$-nearest neighbors \cite{zhong2019invariance,yang2018leveraging}. Among them, clustering based methods have reported very competitive accuracy for UDA in person re-ID. These methods usually employ an iterative process of predicting pseudo identities for unlabeled target samples according to the clusters and fine-tuning the model with those predicted samples. Despite their promising results, clustering based methods are restricted by two main drawbacks. On the one hand, the clustering accuracy can not be guaranteed even using the modern approaches, so that pseudo labels assigned by clusters can be noisy. Training the model with noisy labels that assigned to wrong identities will undoubtedly damage the re-ID performance. On the other hand, most clustering methods tend to leave low confidence samples as outliers and do not assign cluster labels to them, \textit{e.g.}, DBSCAN \cite{ester1996density}. These outliers are usually hard samples that encounter high image variations. Without considering such samples during training, the model may have a problem in discriminating high variation testing samples. However, directly assigning them to the nearest cluster will bring more noisy labels, hindering the retraining of the model. 

Co-Teaching (CT) \cite{han2018co} is a commonly used algorithm for training model with noisy labels, which learns two networks by feeding samples with small losses of one network to another. However, most co-teaching frameworks utilize symmetric inputs for both networks, which do not effectively apply to the context of clustering based cross-domain re-ID. This is because that the training samples with low-confidence commonly have large losses during training. Using symmetric inputs leads the model to always select easy samples and ignore the low-confidence samples within the training mini-batch. As a consequence, the second shortcoming mentioned above will still remain and will lead re-ID model to a local minimum. 

To this end, we choose the state-of-the-art clustering based method proposed in \cite{song2018unsupervised} as our baseline, and propose an asymmetric co-teaching framework to eliminate the negative effects of the above two shortcomings. Specifically, we first divide the target samples into inliers and outliers, according to the clustering results (as shown in Fig.~\ref{fig:process}). In this paper, we regard the low-confident samples recognized by the clustering method as outliers while remaining as inliers. After that, our framework is trained with two models. The first one is the main model which aims to infer samples with small losses from the inliers, while the second one is the collaborator model that estimates samples with small losses from the outliers. The samples inferred/estimated by the certain model are selected for the training of another model. This training process is similar to the traditional co-teaching, except that the inputs of the two models are asymmetric, \textit{i.e.} the data for training the two models comes from two different data flows. In this manner, selecting samples with small losses ensure that the models can be trained with possibly clean data. Moreover, these two models are iteratively promoted by each other. On the one hand, the main model attempts to mine as pure as possible samples from the inliers for maintaining the basic representation of the collaborator model. On the other hand, the collaborator model tries to select as diverse as possible samples from the outliers for further improving the discriminative ability of the main model. Our contributions are summarized in three-fold:
\begin{itemize}
    \item We introduce to employ co-teaching technique for resisting noisy labels generated by clustering in the cross-domain person re-ID. Experiments show that learning with filtered data can consistently improve adaptation accuracy.
    
    \item We divide the unlabeled target data into inliers and outliers and design an asymmetric co-teaching (ACT) framework to make re-ID model see hard samples at the early stage of adaptation. Experiments demonstrate that the asymmetric approach is more effective in handling hard samples than the symmetric one.
    
    \item{Experiments conducted on three large-scale datasets show that our method can apply to various clustering based methods and produces state-of-the-art adaptation accuracy in person re-ID.}
\end{itemize}

\section{Related Work}
\subsection{Cross-Domain Person Re-identification} 

Recent studies in cross-domain re-ID can mainly group into distribution aligning \cite{PTGAN,SPGAN,MMFA,TJAIDL} and clustering-based adaptation \cite{fan2017unsupervised,song2018unsupervised}. Distribution aligning tries to reduce the distribution gap in a common space, which can be further summarized into image-level and attribute-level. For image-level adaptation methods, PT-GAN \cite{PTGAN} uses Cycle-GAN \cite{zhu2017CycleGAN} or Star-GAN \cite{choi2017stargan} to translate the foreground of labeled source images to target camera style for adaptation. Similarly, SPGAN \cite{SPGAN} utilizes Cycle-GAN and additional constraints named ``self-similarity " and ``domain-dissimilarity" for higher accuracy. However, the image-level adaptation algorithms cannot guarantee the identity of generated images, since the generated images still have a large gap compared with real images. For attribute-level adaptation methods, MMFA \cite{MMFA} tries to align the distribution of mid-level semantic attributes between different datasets by minimizing the mean maximum discrepancy (MMD). TJ-AIDL \cite{TJAIDL} leverages a multi-branch network to establish an identity-discriminative and attribute-sensitive feature representation space most optimal for the target domain. These works require attribute annotations of source data, which are hard to obtain in practice. 

Clustering-based adaptation is another straight-forward approach to adapt re-ID model. Fan \textit{et al.} \cite{fan2017unsupervised} use the $k$-means \cite{lloyd1982least} to predict pseudo labels of unlabeled target data for model fine-tuning. However, it is hard to decide correct $k$ value for clustering. Song \textit{et al.} \cite{song2018unsupervised} present a DBSCAN-based adaptation method, which can discover the number of clusters based on the density of features. Although clustering-based methods can achieve high re-ID accuracy for domain adaptation, most of them neglect the wrongly labeled samples in the clustering results and directly use them for training, which cloud have a negative influence on model's performance.

\subsection{Learning with Noisy Labels}

Training deep model on noisy dataset have been widely studied in recent years. Transition matrix \cite{sukhbaatar2014training,patrini2017making}, robust loss function \cite{natarajan2013learning,zhang2018generalized} and CleanNet \cite{lee2018cleannet} are three main efforts for this problem. 

Transition matrix tries to capture the transition probabilities between noisy labels and true labels, based on an assumption that the transition probability between different classes are identical. Sukhbaatar \textit{et al.} \cite{sukhbaatar2014training} add an extra linear layer to capture the relationship between true and corrupted labels. Patrini \textit{et al.} \cite{patrini2017making} estimate the transition matrix by a corresponded loss correction algorithm. However, these kinds of methods do not generalize well on large scale dataset.

Another solution is to design robust loss for model training against noisy labels. Natarajan \textit{et al.} \cite{natarajan2013learning} present an unbiased estimator to give loss correction for model training. Zhang \emph{et al.} \cite{zhang2018generalized} find the drawbacks of mean-absolute loss and cross-entropy loss when applied in this task and further propose a generalized loss function that benefits both losses. However, robust loss methods always have certain constraints, which limit their applications.

Lee \emph{et al.} \cite{lee2018cleannet} propose a CleanNet for tackling this problem. It deploys an additional network to assign a weight score for each sample in training set, and gives lower weight for noise samples to eliminate their negative effects. However, CleanNet needs clean samples for initialization, which can not satisfied for many real-world applications. The co-teaching and co-training frameworks \cite{han2018co,ma2017self} adopt a learning to teach strategy for dealing with noise and unlabeled data. It leverages two networks for synergistic training, in which each network chooses high confidence training samples for the other network. By doing so, these two models can help each other to resist noisy labels. We draw inspiration from these two co-teaching/training methods and develop an asymmetric framework for cross-domain person re-ID.

\begin{algorithm}[!ht]
  \caption{Procedure of the proposed method.}
  \label{alg:asyCo}
  \textbf{Inputs:} Labeled source dataset $\mathcal{S}$, unlabeled target dataset $\mathcal{T}$, 
  ImageNet pre-trained model $M$. Training epochs $e_{1}$, $e_{2}$ and $e_{3}$. Maximum round $r_2$, $r_3$. \\
  \textbf{Outputs:} Adapted model $M_{ada}$. \\
  \begin{algorithmic}[1]
    \STATE ****************** Stage 1 *******************
    \STATE Train $M$ on $\mathcal{S}$ through triplet loss and cross-entropy loss with $e_{1}$ epochs $\Rightarrow$ source model $M_{src}$;
    \STATE ****************** Stage 2 *******************
    \STATE Divide $\mathcal{T}$ into inliers $\mathcal{T}_{i}$ and outliers $\mathcal{T}_{o}$ according 
    to DBSCAN results;
    \STATE Fine-tune $M_{src}$ with $\mathcal{T}_{i}$ for $e_{2}$ epochs and repeat line 4 to 5 for $r_2$ rounds $\Rightarrow$ adapted model $M_{ada}$;
    \STATE ****************** Stage 3 *******************
    \STATE Main model $M_{main}$ $\Leftarrow$ $M_{ada}$, collaborator model $M_{co}$ $\Leftarrow$ $M_{ada}$;
    \FOR{$i$ = 1 to $r_3$ }
      \STATE Deploy $M_{main}$ and DBSCAN to divide $\mathcal{T}$ into $\mathcal{T}_{i}$ and $\mathcal{T}_{o}$;
      \FOR{$j$ = 1 to $e_{3}$}
        \STATE iter = 0;
        \REPEAT
          \STATE Sample mini-batch $t_{i}$ and $t_{o}$ from $\mathcal{T}_{i}$ and $\mathcal{T}_{o}$;
          \IF{iter \% 2 == 0}
            \STATE \textit{// Optimize Main Model.}
            \STATE Deploy $M_{co}$ to choose samples with small loss values from $t_{o}$ and mix them with $t_{i}$ for optimizing $M_{main}$;
          \ELSE 
            \STATE \textit{// Optimize Collaborator Model.} 
            \STATE Deploy $M_{main}$ to choose samples with small loss values from $t_{i}$ for optimizing $M_{co}$;
          \ENDIF
        \STATE iter ++;
        \UNTIL{$\mathcal{T}_{i}$ has been enumerated}
      \ENDFOR
    \ENDFOR
    \STATE $M_{ada}$ $\Leftarrow$ $M_{main}$.
  \end{algorithmic}
\end{algorithm}

\section{The Proposed Method}
\subsection{Overview}

Let $\mathcal{T}$ be the unlabeled target training set and $\mathcal{S}$ be the labeled source training set. Unsupervised domain adaptation tries to leverage both $\mathcal{T}$ and $\mathcal{S}$ to learn a re-ID model that can generalize well on the target testing set. 

Our proposed ACT is designed to solve this problem, which contains three stages: (1) \textit{Source Model Training}. (2) \textit{Clustering-based Adaptation}. (3) \textit{Asymmetric Co-Teaching for Adaptation}. The first two stages aim to obtain a model with basic discriminability by initializing on labeled source data and fine-tuning on target data with pseudo labels generated by the clustering. The third stage attempts to cope with the noisy labels through the cooperation of two models (main model $M_{main}$ and collaborator model $M_{co}$) initialized by the adapted model in the second stage. During the third stage, we first split the target training set into inliers ($T_i$) and outliers ($T_o$) according to the clustering results and then train the model $M_{main}$/$M_{co}$ with  the  small-loss samples  from $T_o$/$T_i$ mined by $M_{co}$/$M_{main}$. The overall procedure of our method is illustrated in Alg.\ref{alg:asyCo}. Next, we will describe the proposed method in detail.

\subsection{Source Model Training} 
In the first stage of our method, we train the source model $M_{src}$ with labeled source dataset $\mathcal{S}$ through the cross-entropy loss and the triplet loss \cite{HermansBeyer2017Arxiv}. Model $M_{src}$ trained on the source data has the basic discriminability for adaptation.

\subsection{Clustering-based Adaptation}
Due to the domain shift between source and target dataset, the obtained source model $M_{src}$ usually can not generalize well on a new dataset. To solve this problem, we follow a robust and reliable adaptation framework \cite{song2018unsupervised}. It adopts $M_{src}$ (source model) to extract pooling-5 features of target images and divide $\mathcal{T}$ into inliers $\mathcal{T}_i$ and outliers $\mathcal{T}_o$ according to the clustering results of DBSCAN. Further training can proceed with the annotated inliers $\mathcal{T}_i$. We present the adaptation algorithm by introducing its distance metric for clustering and loss function.

\textbf{Distance metric for clustering.}
   $k$-reciprocal encoding and Jaccard distance are chosen to be the distance metric for clustering. In detail, we first compute the pair-wise similarity matrix $\mathbf{M}$ by:
    \begin{equation}
    \label{eq:dis}
         M_{i,j}=\left\{
           \begin{array}{ll}
             e^{-||x_i-x_j||^2}, j \in \mathcal{R^*}(i,k) \\
             0,  otherwise 
           \end{array}
         \right.,
    \end{equation} 
    where $M_{i,j}$ is the similarity between sample $i$ and $j$ by using the pooling-5 feature,  $\mathcal{R^*}(i,k)$ is the refined $k$-reciprocal set for sample $i$ which are obtained by adding some specific reliable constrains as mentioned in \cite{zhong2017re}. 

    After obtaining the similarity matrix $\mathbf{M}$, the Jaccard distance $d_{J}(i,j)$ can be computed by: 
    \begin{equation}
        d_{J}(i,j) = 1-\frac{\sum_{k=1}^{N_t}{min(M_{ik},M_{jk})}}{\sum_{k=1}^{N_t}{max(M_{ik},M_{jk})}},
    \end{equation}
    where $N_t$ is the total image number of the target training dataset. To enhance the degree of similarity, each target feature should close to some source features as mentioned in \cite{panareda2017open}, \textit{i.e.}, to minimize:
    \begin{equation}
        d_{W} = 1-e^{||x_i-N_{s}(x_i)||^2},
    \end{equation}
    where $N_{s}(x_i)$ is the nearest neighbor in the source domain for target image $i$. Taking the $d_{J}$ and $d_{W}$ into consideration, the final distance metric for clustering is computed by:
    \begin{equation}
    \label{eq:src}
        d_{i,j} = \lambda [d_{W}(x_i)+d_{W}(x_j)] + (1-\lambda) d_{J}(i,j),
    \end{equation}
    where $\lambda \in [0,1]$ is the balancing factor, and we set it to 0.1 in this study.

    \textbf{Loss function.}	
    Given the computed distance matrix $\mathbf{M}$, we perform DBSCAN on the unlabeled target dataset $\mathcal{T}$ and divide it into inliers $\mathcal{T}_{i}$ and outliers $\mathcal{T}_o$. Each sample in the $\mathcal{T}_{i}$ is assigned to a cluster. Therefore, we can fine-tune $M_{src}$ with pseudo labels of $\mathcal{T}_{i}$ and update the clustering results based on the optimized $M_{src}$ iteratively. We only use triplet loss for the fine-tuning of $M_{src}$. The triplet loss is computed for each batch data by using both pooling-5 and fc-2048 features via:
	\begin{equation}
        \label{eq:triplet}
        L_{tri}=\sum_{i=1}^{N_b}\Big[||x_p-x_a||_2- ||x_n-x_a||_2+m\Big]_{+},
    \end{equation}
	where $N_b$ is the training batch size, $p$ and $n$ are the most dissimilar positive sample and most similar negative sample for anchor image $a$. $x_p$, $x_n$ and $x_a$ denote corresponding features of positive, negative and anchor samples.

After adaptation, we obtain a much better re-ID model $M_{ada}$. However, as mentioned in Sec.~\ref{sec:da}, features extracted by the model are not reliable enough for downstream clustering task, due to the inconsistent distribution of source and target domain. Therefore, clustering results may contain many noisy labels. To further adapt the model against the noisy dataset, we propose ACT in the third stage.

\subsection{Asymmetric Co-Teaching for Adaptation} 
\label{sec:ACT}
Original co-teaching \cite{han2018co} deploys two networks to find possibly clean labels, \textit{i.e.,} small loss samples in the noisy dataset. By sending these samples mined by one network to another for optimization, the influence of incorrect labels can be largely reduced. 
However, co-teaching does not effectively apply to the context of cross-domain re-ID. On the one hand, selected small loss samples are easy for the model to learn and have a limited positive effect for boosting re-ID accuracy. On the other hand, hard samples with high loss values are difficult to be taken into consideration during the co-teaching process, which may limit the diversity of training samples for adaptation. In short, the conventional co-teaching is prone to make re-ID model converge to local minimum, which is not beneficial to train a robust network.

\begin{table*}[!ht]
    \centering
    \footnotesize
    \caption{Comparison with state-of-the-art methods on Market-1501 (M), DukeMTMC-reID (D) and CUHK03 (C). Our proposed algorithm outperforms \textit{image-level} (SP-GAN, PT-GAN), \textit{attribute-level} (TJ-AIDL, MMFA, CFSM),  \textit{clustering-based} (Theory) and \textit{hybrid} (HHL, ECN) methods by a large margin. ``*": reproduced by this paper.}
    \label{tab:cmpDA}
    \resizebox{\linewidth}{!}{
    \begin{tabular}{c|cc|cc|cc|cc|cc|cc} 
        \hline
        \multirow{2}{*}{Source $\rightarrow$ Target} & 
        \multicolumn{2}{c}{M $\rightarrow$ D} \vline & 
        \multicolumn{2}{c}{M $\rightarrow$ C} \vline & 
        \multicolumn{2}{c}{C $\rightarrow$ M} \vline & 
        \multicolumn{2}{c}{C $\rightarrow$ D} \vline & 
        \multicolumn{2}{c}{D $\rightarrow$ M} \vline & 
        \multicolumn{2}{c}{D $\rightarrow$ C} \\
        \cline{2-13}
          & mAP & rank-1 & mAP & rank-1 & mAP & rank-1 & mAP & rank-1 & mAP & rank-1 & mAP & rank-1 \\
        \hline
        Direct Transfer & 14.7 & 28.1 & 10.4 & 12.5 & 19.1 & 40.5 & 7.1 & 16.4 & 19.1 & 46.8 & 11.0 & 12.2 \\
        \hline
        PT-GAN \cite{PTGAN} & - & 27.4 & - & - & - & 31.5 & - & 17.6 & - & 38.6 & - & -\\
        SP-GAN \cite{SPGAN} & 22.3 & 41.1 & - & - & 19.0 & 42.8 & - & - & 22.8 & 51.5 & - & - \\
        TJ-AIDL \cite{TJAIDL} & 23.0 & 44.3 & - & - & - & - & - & - & 26.5 & 58.2 & - & -\\
        MMFA \cite{MMFA} & 24.7 & 45.3 & - & - & - & - & - & - & 27.4 & 56.7 & - & -\\
        CFSM \cite{chang2019disjoint} & 27.3 & 49.8 & - & - & - & - & - & - & 28.3 & 61.2 & - & -\\
        HHL \cite{zhong2018generalizing} & 27.2 & 46.9 & - & - & 29.8 & 56.8 & 23.4 & 42.7 & 31.4 & 62.2 & - & -\\
        ECN \cite{zhong2019invariance} & 40.4 & 63.3 & - & - & - & - & - & - & 43.0 & 75.1 & - & -\\
        Theory* \cite{song2018unsupervised} & 48.4 & 67.0 & 46.4 & 47.0 & 51.2 & 71.4 & 32.2 & 49.4 & 52.0 & 74.1 & 28.8 & 28.5\\
        \hline
        ACT (Ours) & \textbf{54.5} & \textbf{72.4} & \textbf{48.9} & \textbf{49.5} & \textbf{64.1} & \textbf{81.2} & \textbf{35.4} & \textbf{52.8} & \textbf{60.6} & \textbf{80.5} & \textbf{30.0} & \textbf{30.6}\\
        \hline
      \end{tabular}
    }
\end{table*}

To handle the issues mentioned above, we propose a novel co-teaching-like framework for unsupervised cross-domain re-ID in the third stage. In our framework, we initialize main model $M_{main}$ and collaborator model $M_{co}$ with previously obtained $M_{ada}$. $M_{main}$ and $M_{co}$ are then trained in asymmetric manners. $M_{co}$ tries to infer pure data from the outliers for training of the $M_{main}$, which encourages $M_{main}$ to train with more reliable but diverse samples. $M_{main}$ focuses on mining as clean as possible samples from inliers for the training of $M_{co}$, which ensures that $M_{co}$ can maintain the basic representation with easily clustering samples. The whole ACT process has been shown in Fig.~\ref{fig:process}. Specifically, it contains two steps:

\textbf{Step 1: Inliers/Outliers Generation.}We adopt $M_{ada}$ to initialize $M_{main}$ and $M_{co}$ and then extract pooling-5 features for all unlabeled target images for DBSCAN-based pseudo label assignment. DBSCAN is a density-based clustering method, which assigns pseudo labels for samples in high-density area and regards samples in low-density area as outliers. By doing so, $\mathcal{T}$ can be naturally divided into $\mathcal{T}_{i}$ and $\mathcal{T}_{o}$. In \cite{song2018unsupervised}, the obtained $\mathcal{T}_{o}$ are directly discarded. However, we argue that these images are crucial to further boost re-ID accuracy and must be used in a reasonable way. In view of that, we need to give pseudo labels for $\mathcal{T}_{o}$. In our experiments, we assign a pseudo label for each $\mathcal{T}_{o}$ sample according to its nearest neighbor in $\mathcal{T}_{i}$.

\textbf{Step 2: Asymmetric Co-Teaching.} 
In this step, we employ the main model $M_{main}$ and the collaborator model $M_{co}$ for mining useful clean samples from the noisy data and improve the performance cooperatively. Next, we will introduce the training process of $M_{main}$ and $M_{co}$, respectively.

\textit{(a)	For the training of $M_{main}$}, we sample a mini-batch $t_o$ with $B_s$ samples from $\mathcal{T}_o$, and build corresponding $B_s$ triplets. Then, we adopt $M_{co}$ to compute the triplet loss for each triplet and choose $K$\% anchors with the smallest loss values as possibly clean samples. We combine the selected anchors with another $B_s$ samples ($t_i$) obtained from $\mathcal{T}_i$ to form a training mini-batch for the fine-tuning of $M_{main}$. In this part, $M_{co}$ plays a significant role to encourage $M_{main}$ to receive samples as diverse as possible and the re-trained $M_{main}$ is more capable of discriminating hard samples.

\textit{(b)	For the training of $M_{co}$}, we first sample $B_s$ images from $\mathcal{T}_i$ to form $t_i$, and then utilize $M_{main}$ in (a) to mine small loss samples from $t_i$ for optimizing $M_{co}$. In this part, $M_{main}$ mainly focus on ensuring the training samples for $M_{co}$ as pure as possible the re-trained $M_{co}$ can keep basic discriminability for selecting pure samples.

We repeat step 1 and step 2 of ACT for a certain number of rounds to adequately train both models. After training procedure finished, we regard the well-trained $M_{main}$ as our final adapted model for evaluation. It should be noted that we use a different strategy from the original CT to obtain small-loss samples. In our setting, we choose training samples from triplets. Anchors with the smallest $K$\% triplet losses are selected as the reliable samples for fine-tuning. The mechanism behind our selecting strategy is that high-confident anchors are more likely to achieve small losses.

\section{Experiment}
\label{sec:exp}

\subsection{Experimental Settings}
We conduct experiments on three large-scale benchmark datasets: Market-1501 \cite{zheng2015scalable}, DukeMTMC-reID \cite{ristani2016MTMC,zheng2017unlabeled} and CUHK03 \cite{CUHK03}. The mAP and rank-1 accuracy are adopted as evaluation metrics. We use the new-protocol proposed in \cite{zhong2017re} for evaluating CUHK03.

In the \textit{source model training} stage, we adopt both cross-entropy loss and triplet loss for the training of ImageNet-pretrained ResNet-50 model. Adam solver is used to optimize the re-ID model with an initial learning rate of $3 \times 10^{-4}$. We train re-ID model for 150 epochs and the learning rate is linearly decreased to 0 for the last 50 epochs. Margin $m$ in the triplet loss is set to $0.3$. Training batch size $B_s=64$. Input images are resized to $128 \times 64$. We also use random flip and random erasing \cite{RE} for data argumentation.

In the \textit{clustering-based adaptation} stage, we constrain the minimum size of a cluster to 4 and set density radius $p=1.6 \times 10^{-3}$. After a clustering step, we train the model for 30 epochs, and iterate this procedure for 30 rounds. Other parameters are kept the same as in \cite{song2018unsupervised}.

In the last \textit{asymmetric co-teaching} stage, we form triplet samples in a batch to compute triplet loss for each anchor image. Anchors with the smallest $K$\% losses are selected for further training.  We set the small loss ratio $K=20\%$ and linearly increase it to 100\% for the whole $R_{co}$ epochs, $R_{co}=10$.
Adam is used to fine-tune the models for 10 epoch with a fixed learning rate of $6 \times 10^{-5}$.

\subsection{Comparison with state-of-the-arts} 

In Tab.~\ref{tab:cmpDA}, we compare our method with several state-of-the-art jobs on three large scale benchmarks (Market-1501, DukeMTMC-reID and CUHK03). Our method outperforms other algorithms by a large margin on all tasks. Take D$\rightarrow$M and M$\rightarrow$D tasks for example. For image-level SP-GAN, it can only get slightly performance gain than the direct transfer baseline. The attribute-level adaptation methods can achieve better performance than the image-level ones. However, our ACT achieves around 32\% and 27\% mAP improvement over the MMFA and the CFSM. The hybrid method ``ECN" investigates the exemplar-, camera-, and neighborhood-invariance inside the target domain, and can boost the mAP to 43\% and 49\% on both D$\rightarrow$M and M$\rightarrow$D tasks. However, their method still achieves lower re-ID accuracies than ours. Compared with the clustering-based method ``Theory", our ACT can boost mAP scores by 6.9\% and 5.5\% on both adaptation tasks. Similar results can be found in other adaptation tasks. Our method has better accuracies on all experiments, which demonstrates the validity of ACT.

\begin{figure}[!t]
  \centering
  \includegraphics[width=\linewidth]{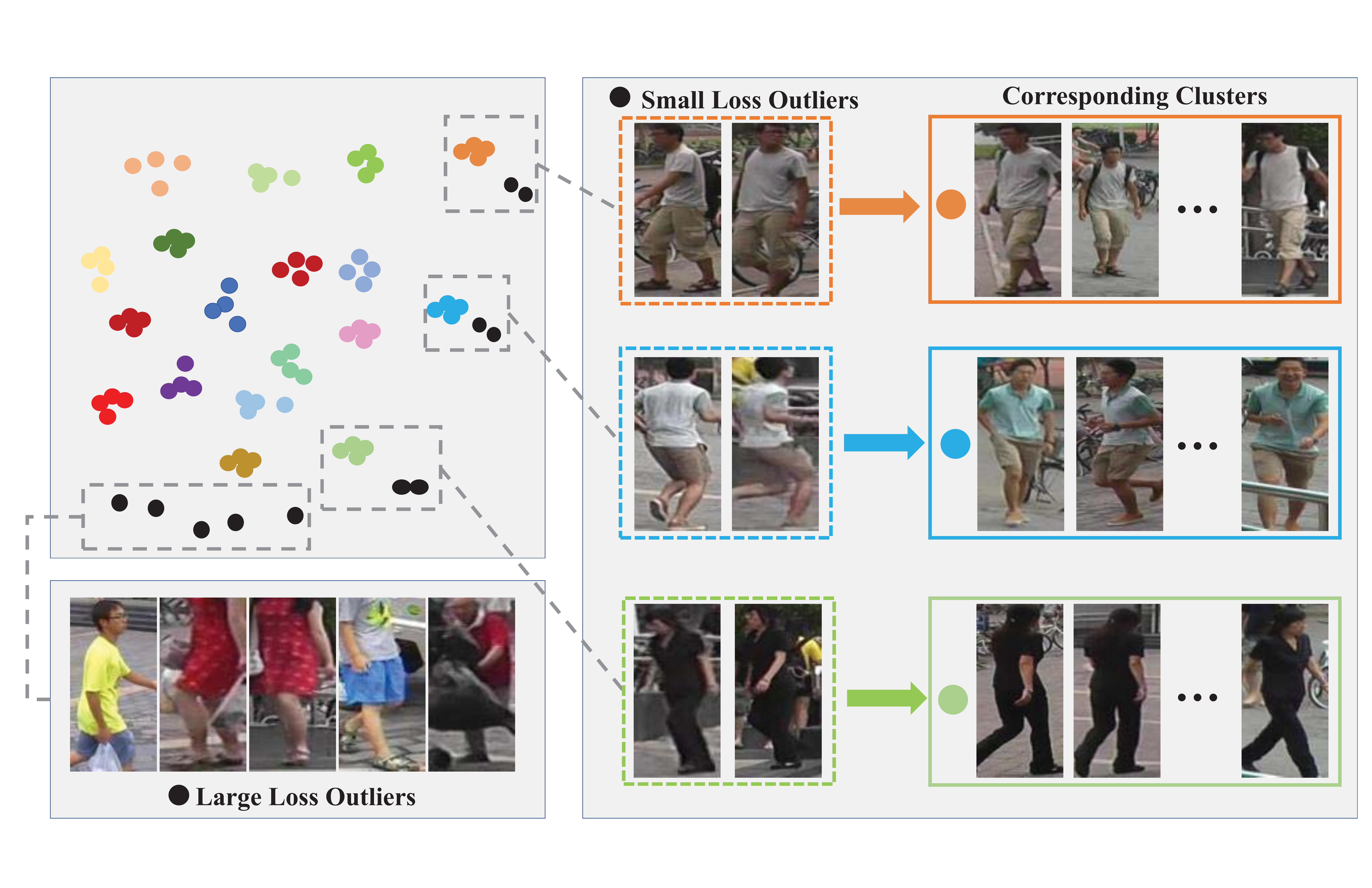}
  \caption{Visualization of small loss samples. We choose images with small loss values from $\mathcal{T}_{i}$ and merge them into their corresponding clusters. Chosen images are not serious affected by illumination and occlusion compared with most images in $\mathcal{T}_{i}$, which may be helpful for model refinement.}
  \label{fig:small}
\end{figure}

\subsection{Visualization of Small Loss Samples} 
In Fig.~\ref{fig:small}, we visualize the small loss samples mining of the collaborator network during the ACT procedure on the Market-1501 dataset. As can be seen, most outliers in $\mathcal{T}_{o}$ are high variance samples caused by occlusion and illumination, which can be hardly assigned correct pseudo labels. After computing the loss in $\mathcal{T}_{i}$ against the collaborator network, we can obtain relative reliable training images in $\mathcal{T}_{i}$ with high diversity, and these samples are helpful for training the main network. 

\begin{figure}[!t]
  \centering
  \includegraphics[width=\linewidth]{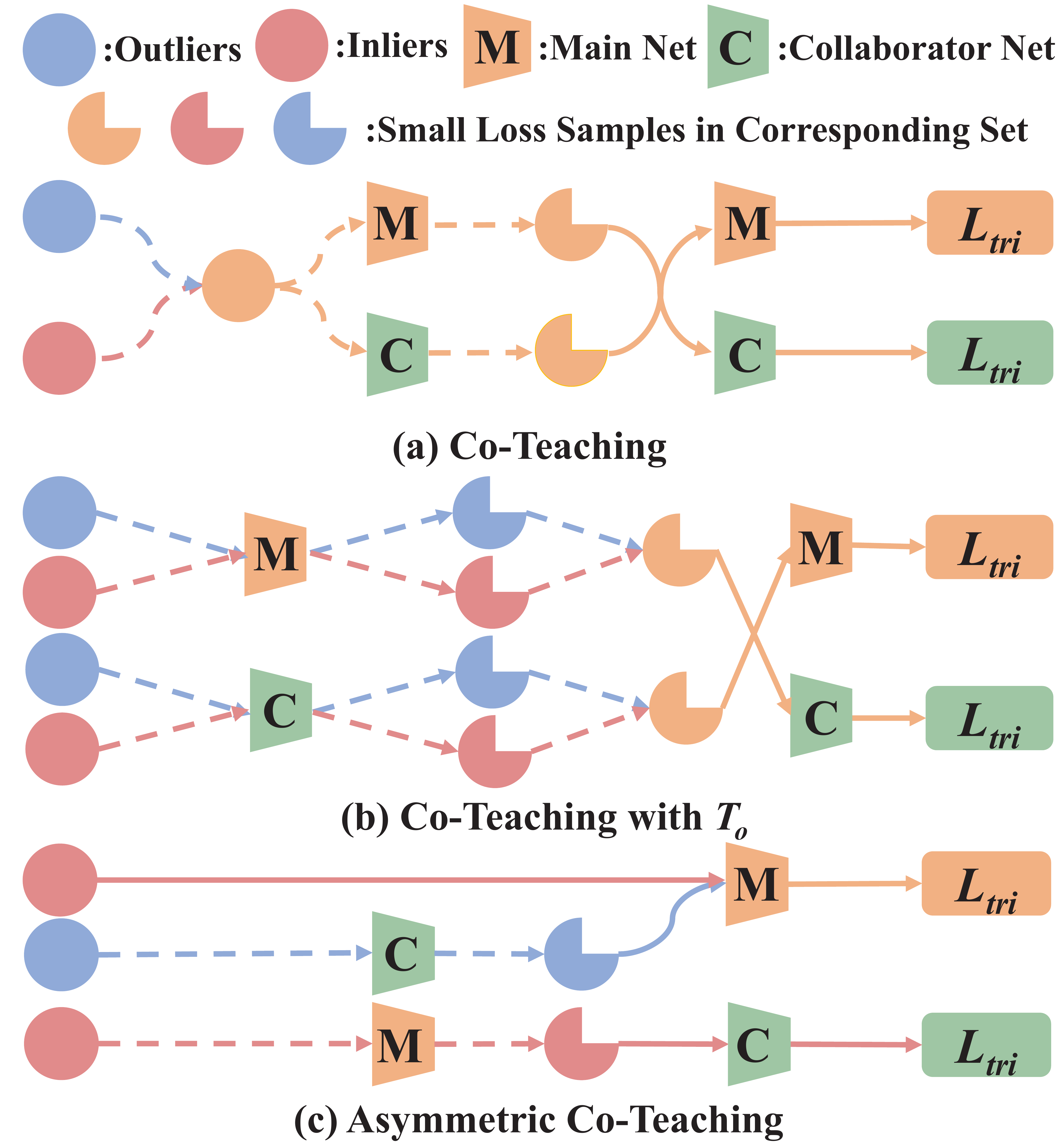}
  \caption{Three different learning strategies compared in our ablation study. We compare \textit{Co-Teaching}, \textit{Revised Co-Teaching (CT with $\mathcal{T}_o$)} and \textit{Asymmetric Co-Teaching} to see which structure achieves the highest re-ID accuracies. Dash lines are basic operations without back-propagation process.}
  \label{fig:structure}
\end{figure}

\begin{table*}[!t]
    \centering
    \caption{Ablation study. We evaluate five settings. \textit{``Theory"}, \textit{``Theory" with outliers}, \textit{co-teaching (CT)}, \textit{CT with outliers} and our \textit{asymmetric co-teaching (ACT)}. Our method gives the best result among other competitors.``*":reproduced by this paper.}
    \begin{tabular}{c|cc|cc|c|c|c} 
        \hline
        \multirow{2}{*}{Methods} & \multicolumn{2}{c}{Duke $\rightarrow$ Market} \vline & \multicolumn{2}{c}{Market $\rightarrow$ Duke} \vline  & \multicolumn{3}{c}{Attributes} \\
        \cline{2-8}
          & mAP & rank-1 & mAP & rank-1 & Clean Label & Hard Sample & Asymmetric Structure \\
        \hline
        Theory* & 52.0 & 74.1 & 48.4 & 67.0 &$\times$&$\times$&$\times$\\
        Theory*+$\mathcal{T}_{o}$ & 54.7 & 76.1 & 49.1 & 68.6 &$\times$&$\surd$&$\times$\\
        CT & 55.0 & 76.4 & 47.3 & 69.8 &$\surd$&$\times$&$\times$\\
        CT + $\mathcal{T}_o$ & 55.3 & 76.8 & 49.6 & 68.9 &$\surd$&$\surd$&$\times$\\
        ACT  & \textbf{60.6} & \textbf{80.5} & \textbf{54.5} & \textbf{72.4} &$\surd$&$\surd$&$\surd$\\
        \hline
      \end{tabular}
    \label{tab:as}
\end{table*}

\subsection{Ablation Study}
Our proposed framework achieves better performance by using asymmetric structure. To argue the effectiveness of our proposed method, we conduct extensive experiments under three different settings in Fig.~\ref{fig:structure} to show: \textbf{(1)} The necessity of taking $\mathcal{T}_o$ into training process. \textbf{(2)} The necessity of clean noisy labels. \textbf{(3)} The effectiveness of asymmetric structure. All the experiments are conducted in D$\rightarrow$M and M$\rightarrow$D tasks. We take ``Theory" \cite{song2018unsupervised} as our baseline model and its re-ID accuracies are shown in the first line of Tab.~\ref{tab:as}.

For \textbf{(1)}, we directly merge $\mathcal{T}_o$ into the training process by giving labels to $\mathcal{T}_{o}$ based on their nearest neighbour in $\mathcal{T}_{i}$. The new dataset is then sent to re-ID network for fine-tuning. As the second line of Tab.~\ref{tab:as} shows, merging $\mathcal{T}_{o}$ into $\mathcal{T}_{i}$ can improve mAP scores by 2.7\% and 0.7\% for both adaptation tasks, which indicates the necessity of taking $\mathcal{T}_{o}$ into training process. However, possible noisy labels in training samples may hinder the further improvement on re-ID accuracies. To demonstrate \textbf{(2)}, we conduct CT to filter out noisy samples. As the third line of Tab.~\ref{tab:as} shows, the original CT has a slight improvement for adaptation in a certain degree. The results may be caused by the aforementioned weakness, so we conduct another experiment, which takes $\mathcal{T}_o$ into each round of CT process to make re-ID model escape the local minimum. The details of CT and revised CT (CT with $\mathcal{T}_{o}$) are shown in Fig.~\ref{fig:structure}-(a) and Fig.~\ref{fig:structure}-(b). From the fourth line of Tab.~\ref{tab:as} we may see that CT with outliers can achieve higher but not significant mAP scores. For \textbf{(3)}, we further evaluate the effectiveness of our proposed asymmetric structure. Detailed workflow in ACT is shown in Fig.~\ref{fig:structure}-(c). Our ACT can achieve the highest accuracies on both adaptation tasks with 60.6\% and 54.5\% mAP scores.

\begin{table}[!t]
    \centering
    \caption{Asymmetric co-teaching with different clustering methods. ACT is compatible with different clustering methods like $k$-means and DBSCAN.}
    \label{tab:clusterAS}
    \begin{tabular}{c|cc|cc} 
        \hline
        \multirow{2}{*}{Methods} & \multicolumn{2}{c}{Duke $\rightarrow$ Market} \vline & \multicolumn{2}{c}{Market $\rightarrow$ Duke}\\
        \cline{2-5}
          & mAP & rank-1 & mAP & rank-1\\
        \hline
        $k$-means & 52.7 & 74.4 & 46.7 & 66.7 \\
        +ACT (20\%) & 56.0 & 76.8 & 49.8 & 69.6 \\
        +ACT (30\%) & 55.0 & 75.5 & 49.5 & 68.1 \\
        \hline
        DBSCAN & 52.0 & 74.1 & 48.4 & 67.0 \\
        +ACT  & 60.6 & 80.5 & 54.5 & 72.4\\
        \hline
      \end{tabular}
\end{table}

\subsection{Variant Evaluation}

\begin{figure}[!t]
  \centering
  \includegraphics[width=\linewidth]{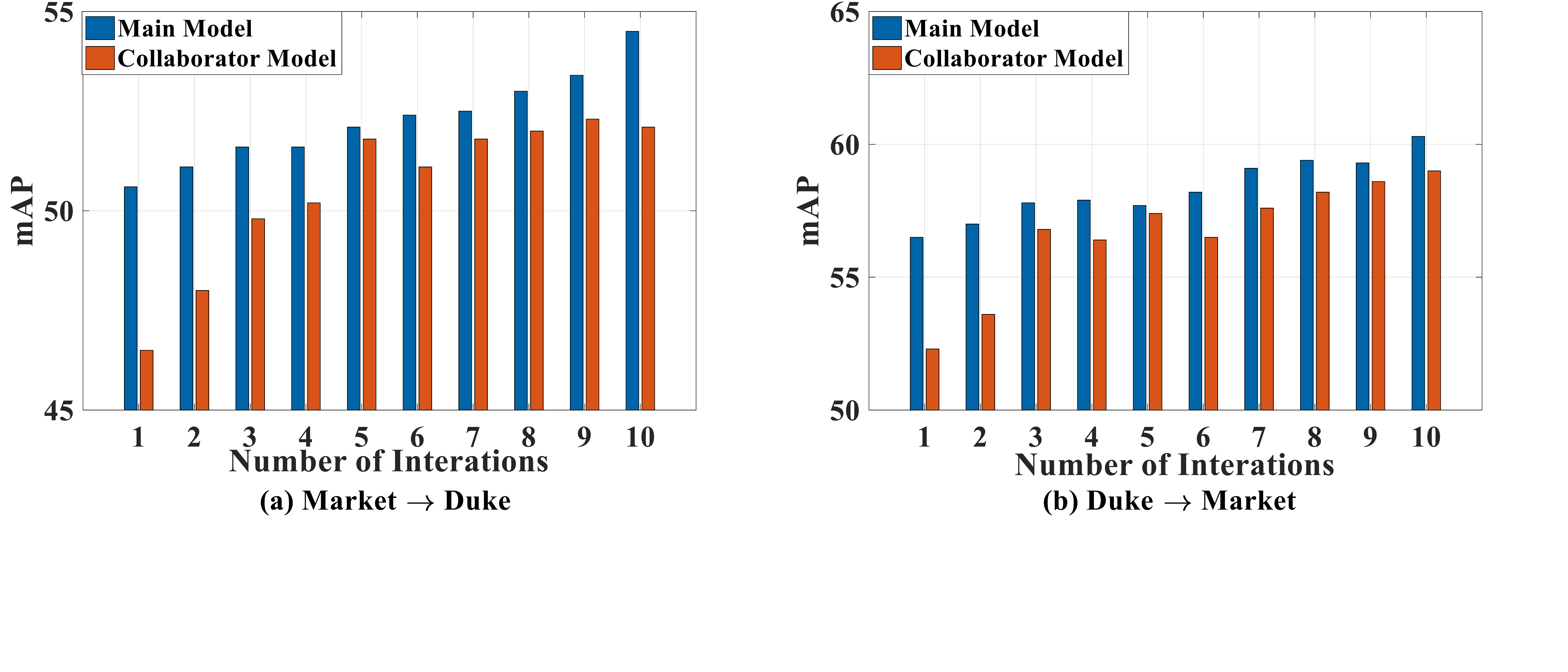}
  \caption{Evaluation of re-ID accuracies for two networks. Main model is always better than collaborator model because of diversity and purity of training data.}
  \label{fig:modelB}
\end{figure}

\textbf{How much difference between the main model and the collaborator model?}
To show the difference between these two models, we report mAP and rank-1 scores of both models during the ACT stage. As can be seen from Fig.~\ref{fig:modelB}, $M_{co}$ is always inferior to $M_{main}$. Since the collaborator model only accepts samples with small loss values mined from the inliers $\mathcal{T}_i$ by $M_{main}$, its training data are more likely to have more easy samples than hard ones.

\begin{figure}[!t]
  \centering
  \includegraphics[width=\linewidth]{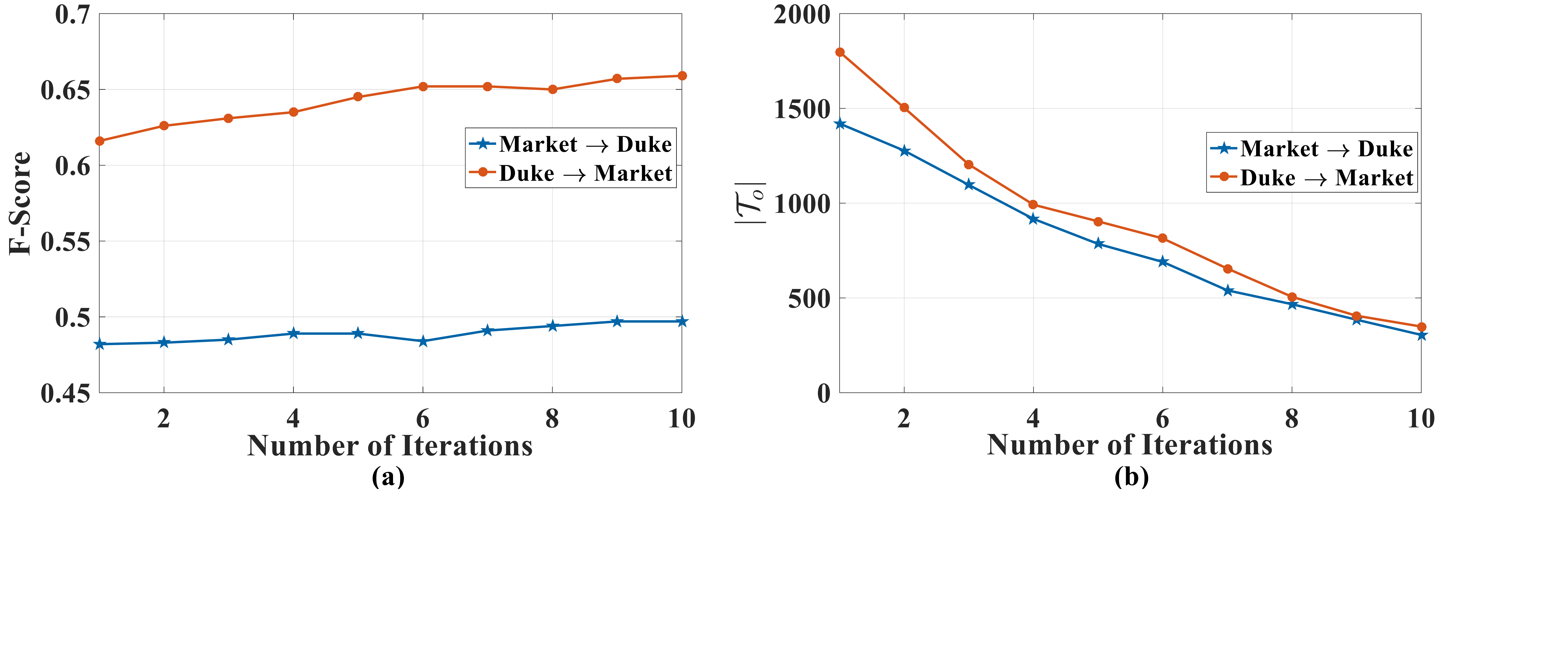}
  \caption{Evaluation of clustering quality during asymmetric co-teaching. (a), F-scores during the adaptation. (b), number of images in $\mathcal{T}_{o}$ ($|\mathcal{T}_{o}|$) for the whole 10 epochs. We may say that asymmetric co-teaching has positive effect for cross-domain re-ID.}
  \label{fig:cluster}
\end{figure}

\textbf{Whether the clustering accuracy increase along with the training iteration?}
We evaluate the clustering results after each DBSCAN clustering step to see whether the accuracy improves. We adopt F-score to measure the quality of clustering after merging $\mathcal{T}_{o}$ images. As shown in Fig.~\ref{fig:cluster}-(a), F-score continues increasing for each iteration, which means the discriminative of the adapted model is also increasing. In Fig.~\ref{fig:cluster}-(b), we also record the size of outliers $\mathcal{T}_{o}$. With the decrease of $\mathcal{T}_{o}$, we can get better clusters for adaptation.


\textbf{Is the proposed pipeline compatible with other clustering methods ?}
We replace the DBSCAN in our method with $k$-means to see whether the proposed pipeline is compatible with other clustering methods. It should be noted that $k$-means does not generate outliers directly, which is different from DBSCAN. However, we can consider the furthest $u\%$ samples far away from their centroids as outliers. In our experiment, we conduct $u$=20 and $u$=30 to see whether ACT benefits $k$-means-based adaptation methods. For convenience, we set $k$ exactly the number of identities for target dataset. As is shown in Tab.~\ref{tab:clusterAS}, when $u$=20, we achieve 56.0\% and 49.8\% mAP scores for both transferring tasks. For $u$=30, our method still outperforms the vanilla $k$-means-based adaptation with 55.0\% and 49.5\% mAP scores, respectively. Our ACT can achieve 3.3\% and 3.1\% improvements on both transferring tasks. In the last two lines of Tab.~\ref{tab:clusterAS}, we compare $k$-means-based algorithm with DBSCAN-based version. The results demonstrate that ACT can also benefit other clustering-based adaptation methods.

\section{Conclusion}
In this paper, we propose a novel asymmetric co-training framework for unsupervised cross-domain re-ID. Our framework is composed of two networks initialized with the same weights named ``Main Model" and ``Collaborator Model". By selecting possibly clean samples from target for each other, adapted main model can resist noisy labels. Furthermore, we design different data flow for both networks to make main model accepts training samples as diverse as possible while collaborator model as pure as possible. The proposed method works fine on three large-scale datasets. We consider applying our work to more unsupervised domain adaptation tasks such as face recognition in the future.

\section*{Acknowledgment}
This work is supported by the National Nature Science Foundation of China (No. 61876159, 61806172, 61572409, U1705286 \& 61571188), the China Postdoctoral Science Foundation Grant (No. 2019M652257), the National Key Research and Development Program of China (No. 2018YFC0831402).

\bibliographystyle{aaai}
\bibliography{aaai}
\end{document}